\documentclass[letterpaper]{article}
\usepackage{authblk}
\usepackage{proceed2e}

\usepackage[margin=1in]{geometry}

\usepackage{times}
\usepackage{fixltx2e}

\usepackage[protrusion=true,expansion=true]{microtype}	
\usepackage{amsmath,amsfonts,amsthm} 
\usepackage{mathtools,mathabx}
\usepackage[mathcal]{euscript}
\usepackage[pdftex]{graphicx}	
\usepackage{url}
\usepackage{booktabs}
\usepackage{rotating}
\usepackage{dsfont}
\usepackage{caption}
\usepackage{changepage}
\usepackage[labelformat=simple]{subcaption}

\usepackage{bm}
\usepackage{siunitx}
\usepackage{nicefrac}
\usepackage{subcaption}
\usepackage{adjustbox}
\usepackage{blindtext}
\usepackage[ruled]{algorithm2e}
\DontPrintSemicolon

\newcommand{\argmin}{\arg\!\min} 
\newcommand{\argmax}{\arg\!\max} 
\newcommand\Dir{\ensuremath{\mathrm{Dir}}} 
\newcommand{\safe}{\emph{safe}}

\newtheorem{definition}{Definition}

\title{Safe Semi-Supervised Learning of Sum-Product Networks}

\author[1,2]{{\bf Martin Trapp}}
\author[2]{{\bf Tamas Madl}}
\author[3]{{\bf Robert Peharz}}
\author[1]{{\bf Franz Pernkopf}}
\author[2]{{\bf Robert Trappl}}
\affil[1]{Signal Processing and Speech Communication Laboratory, Graz University of Technology, Graz, Austria}
\affil[2]{Austrian Research Institute for Artificial Intelligence, Vienna, Austria}
\affil[3]{Computational and Biological Learning Lab, University of Cambridge, Cambridge, UK} 

\begin{document}

\maketitle

\begin{abstract}
In several domains obtaining class annotations is expensive while at the same time unlabelled data are abundant. 
While most semi-supervised approaches enforce restrictive assumptions on the data distribution, recent work has managed to learn semi-supervised models in a non-restrictive regime.
However, so far such approaches have only been proposed for linear models.
In this work, we introduce semi-supervised parameter learning for Sum-Product Networks (SPNs).
SPNs are deep probabilistic models admitting inference in linear time in number of network edges.
Our approach has several advantages, as it (1) allows generative and discriminative semi-supervised learning, (2) guarantees that adding unlabelled data can increase, but not degrade, the performance (\safe{}), and (3) is computationally efficient and does not enforce restrictive assumptions on the data distribution.
We show on a variety of data sets that \safe{} semi-supervised learning with SPNs is competitive compared to state-of-the-art and can lead to a better generative and discriminative objective value than a purely supervised approach. 
\end{abstract}

\section{\uppercase{Introduction}}
In several domains, unlabelled observations are abundant and cheap to acquire, while obtaining class labels is expensive and sometimes infeasible for large amounts of data. In such cases, semi-supervised learning can be used to exploit large amounts of unlabelled data in addition to labelled data. Examples include text \cite{Yang2016} or image data \cite{Kingma2014,Rasmus2015,Maaloe2016}, which are ubiquitous online, but also biological (genomics, proteomics, gene expression) data \cite{Zhang2015} and speech \cite{Thomas2013}.

One of the challenges facing most semi-supervised learning approaches is scalability, many methods scale quadratically or even cubically with data set size, or require restrictive assumptions such as low dimensionality or sparsity \cite{Zhu2009,Kingma2014}. 
In fact, if the data violates the assumptions enforced by a learner, the use of additional unlabelled data can even degrade the classification performance.

Several approaches for semi-supervised have been proposed, including self-training, Transductive Support-Vector Machines (TSVM)~\cite{Collobert2006}, and graph-based methods. We refer to \cite{Zhu2009,Hady2013} for comprehensive reviews on the state-of-the-art. As pointed out by \cite{Kingma2014}, self-training is error-prone (it can reinforce poor predictions) and TSVM as well as graph-based methods are difficult to scale. In addition, TSVM, and its recent extensions \cite{Li2015} require that the decision boundary lie in a low density region, yielding sub-optimal accuracy if this is not met. Each of these methods can lead to decreased accuracy when adding unlabelled data. 
To overcome these limitations, \cite{Loog2015} recently proposed a probabilistic formulation for \safe{} semi-supervised learning of generative linear models.

In the family of deep probabilistic models, Sum-Product Networks (SPNs) \cite{Poon2011} have recently gained popularity, due to their efficiency, i.e. linear-time inference, generality, i.e. they subsume existing approaches such as latent tree models and mixtures, and performance on various tasks including computer vision \cite{Poon2011,Gens2012}, action recognition \cite{Amer2012}, speech \cite{Peharz2014}, and language modelling \cite{Cheng2014}.

For probabilistic models, including SPNs, semi-supervised learning with generative models is natural. 
Data points are assigned to whichever class maximizes $p(\bm x, y)=p(y)p(\bm x|y)$, with $p(\bm x|y)$ being a generative model for the data in class $y$. Subsequently, the labelled data points can be used to learn the model. 
Unfortunately, adding unlabelled data can significantly degrade classification accuracy instead of improving it \cite{Cozman2002}. 

In this paper, we introduce \safe{} semi-supervised parameter learning for SPNs that is \safe{}, scalable and non-restrictive. 
\emph{Safe} means that adding unlabelled data can increase, but not degrade, model performance.
The training time scales linearly with added data points and apart from the structure of the underlying SPN, no assumptions are made regarding the data distribution. Unlike other semi-supervised methods, the presented approach does not need low-density or clustering assumptions \cite{Chapelle2006}. In addition to safety, we show competitive results when compared with state-of-the-art approaches in Section 4.

The structure of the paper is as follows: 
Section \ref{sec:background} introduces the notation used throughout the paper, describes recent approaches for parameter learning in SPNs and introduces the contrastive pessimistic likelihood estimation for \safe{} semi-supervised learning of generative models. 
In Section \ref{sec:SSLSPNs} we propose \safe{} semi-supervised learning for SPNs, give derivations for generative and discriminative parameter learning and present the algorithm MCP-SPN for training \safe{} semi-supervised SPNs.
Experiments are presented in Section \ref{sec:Experiments} showing that \safe{} semi-supervised SPNs are able to escape from degenerated supervised solutions, generally outperform purely supervised learning and achieve competitive performance on a variety of data sets.
Section \ref{sec:Conclusion} concludes the paper and gives future prospects.

\section{\uppercase{Background}}\label{sec:background}
We use capital letters to denote random variables (RVs) and denote a set of RVs as $\bm X = \{X^d\}_{d=1}^D$. 
Moreover, we denote a realisation of a RV using lower-case letters and indicate a realisation of $\bm X$ using bold lower-case letters, e.g. $\bm x = \{x^d\}_{d=1}^D$.
We denote the set of labelled observation as $\mathcal{X} = \{(\bm x_n, \bm y_n)\}_{n=1}^N$ and the set of unlabelled observation as $\mathcal{U} = \{\bm u_m\}_{m=1}^M$ where $\bm x_n$, $\bm u_m$ are the features and $\bm y_n$ the labels in one-hot-encoding. 
Additionally, we use $\bm q = \{\bm q_m\}_{m=1}^M$ to denote soft labels for the unlabelled observations.
We generally write $p(x)$ instead of $p(X = x)$ and write $p(\bm x)$ instead of $p(\bm X = \bm x)$. 
For readability, we will refer to the value of an SPN using a calligraphic notation, $\mathcal{S}[\bm x]$, and write $S_i[\bm x]$ for the value of the $i$\textsubscript{th} node in an SPN.

\subsection{\uppercase{Sum-Product Networks}}
SPNs are a deep probabilistic architecture which allows to capture expressive variable interactions, yet guaranteeing exact computations of marginals in linear time. 
SPNs have its foundation in network polynomials for efficient inference in Bayesian networks introduced by \cite{Darwiche2009}. 
Poon and Domingos~\cite{Poon2011} generalized the idea and introduced SPNs over random variables (RVs) with finitely many states.

\pagebreak
\begin{definition}
(Sum-Product Network \cite{Poon2011})
A sum-product network (SPN) over variables $X^1, \dots, X^d$ is a rooted directed acyclic graph whose leaves are the indicators $x^1, \dots, x^d$ and $\bar{x}^1, \dots, \bar{x}^d$ and whose internal nodes are sums and products. Each edge $(i,j)$ emanating from a sum node $i$ has a non-negative weight $w_{ij}$. The value of a product node is the product of the values of its children. The value of a sum node is $\sum_{j \in Ch(i)} w_{ij} v_j$, where $Ch(i)$ are the children of $i$ and $v_j$ is the value of node $j$. The value of an SPN
$\mathcal{S}[x^1, \bar{x}^1, \dots, x^d, \bar{x}^d]$ is the value of its root.
\end{definition}
SPNs can be generalized by replacing the leaf node indicators with arbitrary input distributions~\cite{Peharz2015b}. 
Thus, we consider SPNs with arbitrary leaf node distributions throughout the paper.

\subsubsection{Generative Learning}
The parameters of an SPN can be learned efficiently using Expectation Maximisation (EM)~\cite{Poon2011,Peharz2016}. 
We use the formulation of \cite{Peharz2016}, where the updates for the parameters of the $i$\textsubscript{th} sum node are defined as:
\begin{align} 
    n_{ij} =& w_{ij} \sum_{n = 1}^N \frac{1}{\mathcal{S}[\bm x_n]} \frac{\partial \mathcal{S}[\bm x_n]}{\partial S_i} S_j[\bm x_n] \text{, and}  \\
    w_{ij} \leftarrow& \frac{n_{ij}}{\sum_{l \in Ch(i)} n_{il}} \text{.} \label{eq:generativeWeightUpdates}
\end{align}
Furthermore, the parameter update for an exponential family leaf node with scope $d$ and parameter $\theta$ is given by the expected sufficient statistic and can be computed as:
\begin{align} \label{eq:generativeLeaveUpdates}
    g_i(\bm x) =& \frac{1}{\mathcal{S}[\bm x]} \frac{\partial \mathcal{S}[\bm x]}{\partial S_i} S_i[\bm x] \text{, and} \\
    \theta_{i} \leftarrow& \frac{\sum_{n = 1}^N g_{i}(\bm x_n) t(x_n^d)}{\sum_{n=1}^N g_{i}(\bm x_n)} \text{,}
\end{align}
where $t(x)$ denotes the sufficient statistics.
We assume complete evidence for the RVs $\bm X$ and refer to \cite{Peharz2016} for a derivation of the updates with partial evidence.

\subsubsection{Discriminative Learning}
The parameters of a discriminative SPN can be learned by optimising the conditional log likelihood using back-propagation~\cite{Gens2012}. 
The set of variables of a discriminative SPN are divided into query variables $\bm Y$, hidden variables $\bm H$ and observed RVs $\bm X$. 
Therefore, the value of a discriminative SPN is denoted as $\mathcal{S}[\bm Y = \bm y, \bm{H} = \bm h | \bm{X} = \bm x]$. 
Furthermore, the conditional probability is estimated by setting all indicator functions of the hidden variables to $\bm 1$ and computing
\begin{align}
    p(\bm y | \bm x) = \frac{\mathcal{S}[\bm Y = \bm y, \bm H = \bm{1} | \bm X = \bm{x}]}{\mathcal{S}[\bm Y = \bm 1, \bm H = \bm{1} | \bm X = \bm{x}]},
\end{align}
where setting the indicators of the hidden variables to one allows the gradients of the conditional log likelihood to be computed in a single upward pass. 
For the sake of readability, we omit the hidden variables if their indicators are set to one and write $\mathcal{S}[\bm y | \bm x]$ for the value of a discriminative SPN instead.

Given a network structure, one can train a discriminative SPN by gradient ascent using the partial derivatives of the SPN with respect to the parameters of the network.
The partial derivatives of the weights take the form
\begin{align} \label{eq:discriminativeWeightUpdates} 
    \frac{\partial \log p(\bm y | \bm{x})}{\partial w_{ij}} = \frac{1}{\mathcal{S}[\bm y | \bm{x}]} \frac{\partial \mathcal{S}[\bm y | \bm{x}]}{\partial w_{ij}} - \frac{1}{\mathcal{S}[\bm{1} | \bm{x}]} \frac{\partial \mathcal{S}[\bm{1} | \bm{x}]}{\partial w_{ij}} \text{, }
\end{align}
where $\frac{\partial \mathcal{S}}{\partial w_{ij}} = \frac{\partial \mathcal{S}}{\partial S_i} S_j$ is computed using back-propagation.
By setting the gradient of the root node $\frac{\partial \mathcal{S}}{\partial S} = 1$, the gradients of the subsequent nodes are computed in a top-down order. 
At sum nodes the gradient is propagated to the children using $\frac{\partial \mathcal{S}}{\partial S_j} \leftarrow \frac{\partial \mathcal{S}}{\partial S_j} + w_{ij} \frac{\partial \mathcal{S}}{\partial S_i}$ and at product nodes using $\frac{\partial \mathcal{S}}{S_j} \leftarrow \frac{\partial \mathcal{S}}{S_j} + \frac{\partial \mathcal{S}}{\partial S_i} \prod_{l \in Ch(i)\setminus\{j\}} S_l$. 
As indicated, the gradient at a node $j$ is accumulated based on the parents gradients.
We refer to \cite{Gens2012} for further details on the derivation of the gradients and derivations of \emph{hard} gradient updates.

As in network polynomials for Bayesian networks \cite{Darwiche2009}, partial derivatives of any parameter in an SPN can be calculated using the chain rule, leading to a straight forward computation of parameter updates for the leaf node distributions, i.e.
\begin{align}
    \frac{\partial \mathcal{S}[\bm y | \bm{x}]}{\partial \theta} &= \frac{\partial \mathcal{S}[\bm y | \bm{x}]}{\partial S_i} \frac{\partial p(x^d | \theta)}{\partial \theta} \text{, and} \\
     \frac{\partial \log p(\bm y | \bm{x})}{\partial \theta} &= \frac{1}{\mathcal{S}[\bm y | \bm{x}]} \frac{\partial \mathcal{S}[\bm y | \bm{x}]}{\partial \theta} - \frac{1}{\mathcal{S}[\bm{1} | \bm{x}]} \frac{\partial \mathcal{S}[\bm{1} | \bm{x}]}{\partial \theta} \text{.}
\end{align}
In the case of univariate Gaussian distributions, the updates are computed by taking the partial derivatives of the mean and the variance of the distribution.

\subsection{\uppercase{Contrastive Pessimistic Likelihood Estimation}}
Most semi-supervised learning approaches require strong assumptions, e.g. low density assumption for TSVM, and can lead to decreased performance with increasing number of unlabelled data samples if these assumptions are violated. Loog \cite{Loog2015} has proposed Contrastive Pessimistic Likelihood Estimation (CPLE) in order to facilitate performance guarantees while only relying on the assumptions of an underlying generative model. 

CPLE maintains soft labels (hypotheses) for each unlabelled data point, and assigns them pessimistically, using a training objective that maximizes the log likelihood on the data $L(\theta |\cdot)$ but minimizes the improvement provided by the unlabelled data. Therefore, CPLE yields in a \safe{} semi-supervised objective. 

Model parameters under CPLE are estimated according to:
\begin{align}
    \theta^\ast = \argmax_{\theta \in \Theta} \argmin_{\bm q \in \Delta_{K-1}^M} L(\theta | \mathcal{X}, \mathcal{U}, \bm{q}) - L(\theta^+ | \mathcal{X}, \mathcal{U}, \bm q) \text{,}
\label{cple}
\end{align}
where $\bm q$ denotes soft labels for every unlabelled data point and $\theta^+$ denotes the parameters of a purely supervised model derived only on $\mathcal{X}$. The introduction of soft labels, respects the fact that classes may be overlapping. In the case of $K$ unique class labels each soft label vector $\bm q_m$ is an element of the $K-1$ simplex $\Delta_{K-1}$.

Since the trained classifier assumes the worst-case improvement, its performance cannot degrade when adding unlabelled data.
Loog \cite{Loog2015} constrains the CPLE to generative models, and provides a concrete solution for a simple linear classifier based on linear discriminative analysis.
In the following, we define a contrastive pessimistic objective for generative and discriminative SPNs, yielding in a \safe{} semi-supervised learning procedure with linear computational complexity which only relies on the assumptions intrinsic to the given network structure. 

\section{\uppercase{Safe Semi-Supervised SPNs}} \label{sec:SSLSPNs}
Given an SPN $\mathcal{S}[\bm x, \bm y]$ we can find the optimal parameters for generative \safe{} semi-supervised learning using the CPLE objective defined in Equation~\eqref{cple}.
For clarity, we always use the plus operator to indicate parameters of the purely supervised solution, e.g. weights $w^+$, and indicate parameters of the \safe{} semi-supervised solution using an asterisk.
Due to the conservative choice of $\bm q$ by minimizing the improvement over the supervised result, and since we can always take $\theta^\ast = \theta^+$ in the worst case, this objective is guaranteed to lead to a \safe{} solution. 
More formally, as shown in Loog \cite{Loog2015}, it is guaranteed that
\begin{equation}
L( \theta^\ast | \mathcal{X}, \mathcal{U}, \bm{q}) \geq L(\theta^+ | \mathcal{X}, \mathcal{U}, \bm{q}) \text{.}
\end{equation}
Therefore, if log likelihoods are used in the CPLE objective the \safe{} semi-supervised solution has at least the same log likelihood given $\mathcal{X}, \mathcal{U}$ and $\bm{q}$ as the purely supervised objective.

\subsection{\uppercase{Generative Safe Semi-Supervised Learning}}
In the following we derive the Expectation Maximisation (EM) updates for the generative \safe{} semi-supervised SPN.
Therefore, let 
\begin{align}
    \mathcal{S}[\bm x, \bm y | \theta] = \sum_{k=1}^K \mathds{1}_{y_k} w_k S_k[\bm x | \bm y, \theta]
\end{align}
be the likelihood of a semi-supervised SPN for labelled observations $(\bm x, \bm y) \in \mathcal{X}$. 
We denote $\mathds{1}_{y_k}$ to be the indicator for class $k$ which is one if $y_k$ is true and zero otherwise. 
Furthermore, let
\begin{align}
    \mathcal{S}[\bm u, \bm q | \theta] = \sum_{k=1}^K q_k w_{k} S_k[\bm u | \bm q, \theta]
\end{align}
be the likelihood of a semi-supervised SPN for unlabelled observations $(\bm u, \bm q)$ with $\bm q$ being the soft labels of the data.
Note that $\sum_k q_k = 1$ for all unlabelled observations, as each soft label vector is an element of the $K-1$ simplex.
We can therefore define the generative log likelihood function of a semi-supervised SPN as the sum of the log likelihood given the labelled data and the unlabelled data. 
Formally, we define the generative log likelihood of a semi-supervised SPN as
\begin{align}
\begin{split}
L(\theta | \mathcal{X}, \mathcal{U}, \bm{q}) =& \sum_{n=1}^{N} \log \mathcal{S}[\bm x_n, \bm y_n | \theta]\\
     +& \sum_{m=1}^{M} \log \mathcal{S}[\bm u_m, \bm q_m | \theta]
\end{split}
\end{align}
which allows for straightforward derivation of the EM updates. 
The updates of the weights of sum node $S_i$ in $\mathcal{S}$ can be computed as in Eq.~\eqref{eq:generativeWeightUpdates} using the following $n_{ij}$, i.e.
\begin{align}
    \begin{split}
        n_{ij} ={}& w_{ij} \sum_{n=1}^{N} \frac{1}{\mathcal{S}[\bm x_n, \bm y_n]} \frac{\partial \mathcal{S}[\bm x_n, \bm y_n]}{\partial S_i} S_j[\bm x_n | \bm y_n] 
         \\ +& w_{ij} \sum_{m=1}^{M} \frac{1}{\mathcal{S}[\bm u_m, \bm q_m]} \frac{\partial \mathcal{S}[\bm u_m, \bm q_m]}{\partial S_i} S_j[\bm u_m | \bm q_m] \text{, }
    \end{split}\label{eq:generativeSSLWeights}
\end{align}
where we omitted the parametrization of the network for better readability. 
Furthermore, we can update the parameters of an exponential family leaf node with scope $d$ using the expected sufficient statistics as
\begin{align}
    g_{i}(\bm x_n, \bm y_n) =& \frac{1}{\mathcal{S}[\bm x_n, \bm y_n]} \frac{\partial \mathcal{S}[\bm x_n, \bm y_n]}{\partial S_i} S_i[\bm x_n | \bm y_n] \text{,} \\
    g_{i}(\bm u_m, \bm q_m) =& \frac{1}{\mathcal{S}[\bm u_m, \bm q_m]} \frac{\partial \mathcal{S}[\bm u_m, \bm q_m]}{\partial S_i} S_i[\bm u_m | \bm q_m] \text{,}
\end{align}
\begin{align}
    \theta_{i} \leftarrow \frac{\sum_{n=1}^N g_{i}(\bm x_n, \bm y_n) t(x_n^d) + \sum_{m=1}^M g_{i}(\bm u_m, \bm q_m) t(u_m^d)}{\sum_{n=1}^N g_{i}(\bm x_n, \bm y_n) + \sum_{m=1}^M g_{i}(\bm u_m, \bm q_m)} \text{, } \label{eq:generativeSSLParam}
\end{align}
where we assume complete evidence for the RVs $\bm X$ and $\bm U$. 

Subsequently, the soft label for class $k$ of an unlabelled sample $m$ is updated pessimistically with gradient descent using the partial derivative of $q_{mk}$ which is defined as
\begin{align}
    \frac{\partial L(\theta | \mathcal{X}, \mathcal{U}, \bm{q})}{\partial q_{mk}} &= \frac{w_{k} S_k[\bm u_m | \bm q_m, \theta]}{\mathcal{S}[\bm u_m, \bm q_m | \theta]} \text{, and}\\
    \nabla q_{mk} &= \frac{\partial L(\theta^\ast | \mathcal{X}, \mathcal{U}, \bm{q})}{\partial q_{mk}} - \frac{\partial L(\theta^+ | \mathcal{X}, \mathcal{U}, \bm{q})}{\partial q_{mk}}\text{.} \label{eq:slgradient}
\end{align}
Note that after each gradient update it is necessary to ensure that the soft labels for the unlabelled data points are on the $K-1$ simplex. For this purpose, the soft labels are projected back to the $K-1$ simplex using the approach by Duchi et al.~\cite{Duchi2008}.

\subsection{\uppercase{Discriminative Safe Semi-Supervised Learning}}
Conditional likelihoods instead of generative objectives are a more natural way of learning SPNs for classification tasks in the semi-supervised regime. 
Formally, the model parameters for discriminative \safe{} semi-supervised SPNs are estimated according to
\begin{align}
\begin{split}
    \argmax_{\theta \in \Theta}\argmin_{\bm q \in \Delta_{K-1}^M} CL(\theta | \mathcal{X}, \mathcal{U}, \bm q) - CL(\theta^+ | \mathcal{X}, \mathcal{U}, \bm q),
\label{cple_dspn}
\end{split}
\end{align}
where we intentionally use $CL(\theta | \cdot)$ to indicate the use of the conditional log likelihood. 
Extending the formulation for discriminative SPNs allows to define a discriminative learning approach for \safe{} semi-supervised SPNs, i.e.,
\begin{align}
\begin{split}
    CL(\theta | \mathcal{X}, \mathcal{U}, \bm q) =& \sum_{n=1}^{N} \log \mathcal{S}[\bm y_n | \bm x_n, \theta] \\
    +& \sum_{m=1}^{M} \log \mathcal{S}[\bm q_m | \bm u_m, \theta] \text{,}
    \label{eq_cllh}
\end{split}
\end{align}
where the conditional likelihood for labelled and unlabelled data, respectively, are given as
\begin{align}
    \mathcal{S}[\bm y | \bm x, \theta] =& \frac{S_k[\bm x, \bm y | \theta]}{S_k[\bm x, \bm 1 | \theta]} \text{, and} \\
    \mathcal{S}[\bm q | \bm u, \theta] =& \frac{S_k[\bm u, \bm q | \theta]}{S_k[\bm u, \bm 1 | \theta]} \text{.}
\end{align}

The partial derivatives for the weights of the discriminative semi-supervised SPN therefore become
\begin{align}\label{eq:discriminativeSSLWeights}
    \begin{split}
    \frac{\partial}{\partial w_{ij}} &CL(\theta | \mathcal{X}, \mathcal{U}, \bm q) = \\ &\sum_{n=1}^{N}  \frac{1}{\mathcal{S}[\bm y_n | \bm x_n]} \frac{\partial \mathcal{S}[\bm y_n | \bm x_n]}{\partial w_{ij}}
    - \frac{1}{\mathcal{S}[\bm 1 | \bm x_n]} \frac{\partial \mathcal{S}[\bm 1 | \bm x_n]}{\partial w_{ij}} \\
    + &\sum_{m=1}^{M} \frac{1}{\mathcal{S}[\bm q_m | \bm u_m]} \frac{\partial \mathcal{S}[\bm q_m | \bm u_m]}{\partial w_{ij}} - \frac{1}{\mathcal{S}[\bm 1 | \bm u_m]} \frac{\partial \mathcal{S}[\bm 1 | \bm u_m]}{\partial w_{ij}} \text{.}
    \end{split}
\end{align}
Similarly, we can derive the partial derivatives of the leaf node parameters by applying the chain rule, leading to the following parameter updates
\begin{align}
\frac{\partial}{\partial \theta} &CL(\theta | \mathcal{X}, \mathcal{U}, \bm q) =\\
&\sum_{n=1}^{N}  \frac{1}{\mathcal{S}[\bm y_n | \bm x_n]} \frac{\partial \mathcal{S}[\bm y_n | \bm x_n]}{\partial \theta}
    - \frac{1}{\mathcal{S}[\bm 1 | \bm x_n]} \frac{\partial \mathcal{S}[\bm 1 | \bm x_n]}{\partial \theta} \\
    + &\sum_{m=1}^{M} \frac{1}{\mathcal{S}[\bm q_m | \bm u_m]} \frac{\partial \mathcal{S}[\bm q_m | \bm u_m]}{\partial \theta} - \frac{1}{\mathcal{S}[\bm 1 | \bm u_m]} \frac{\partial \mathcal{S}[\bm 1 | \bm u_m]}{\partial \theta} \label{eq:discriminativeSSLParam} \text{.}
\end{align}
To pessimistically update the soft labels, one can use gradient descent on the partial derivatives similar as for the generative objective in Eq.~\eqref{eq:slgradient}.

\begin{algorithm}[!h]
\DontPrintSemicolon
\SetCommentSty{emph}

\KwIn{A valid SPN structure $\mathcal{S}$, labelled data $\mathcal{X}$, unlabelled data $\mathcal{U}$.}
\KwOut{Learned parameters and soft labels.}

\tcp{learn purely supervised SPN}
\eIf{generative}{
    $\theta^+ \gets \argmax_{\theta \in \Theta} \log \mathcal{S}[\bm{x}, \bm{y} | \theta]$
}{
    $\theta^+ \gets \argmax_{\theta \in \Theta} \log \mathcal{S}[\bm{y} | \bm{x}, \theta]$
}    
\tcp{initialize soft labels}
\eIf{optimistic}{
    \ForEach{$k \in \{1, \dots, K\}$}
    {
        $\bm q_k \gets \frac{S_k[\bm u | \theta^+]}{\mathcal{S}[\bm 1 | \bm u, \theta^+]}$
    }
}{
    $\bm q \sim \Dir(\frac{1}{K}, \dots, \frac{1}{K})$
} 
\tcp{learn \safe{} semi-supervised SPN}
\Repeat{convergence or early stopping}{
\tcp{optimistic parameter learning}
\eIf{generative}{
    \tcp{Eq.~\ref{eq:generativeSSLWeights} and Eq.~\ref{eq:generativeSSLParam}}
    $\theta^\ast \gets \argmax_{\theta \in \Theta} \log \mathcal{S}[\bm{x}, \bm{y} | \theta] + \log \mathcal{S}[\bm{u}, \bm{q} | \theta]$
}{
    \tcp{Eq.~\ref{eq:discriminativeSSLWeights} and Eq.~\ref{eq:discriminativeSSLParam}}
    $\theta^\ast \gets \argmax_{\theta \in \Theta} \log \mathcal{S}[\bm{y} | \bm{x}, \theta] + \log \mathcal{S}[\bm{q} | \bm{u}, \theta]$
}  
\tcp{pessimistic soft label adjustment}
$\bm q \gets \bm q - \alpha \nabla \bm q$ \tcp*{Eq.~\ref{eq:slgradient}}
$\bm q \gets$ projectOnSimplex($\bm q$, $\Delta_{K-1}^M$)
}

\Return{$\theta^\ast$ and $\bm q$}
\caption{MCP-SPN}
\label{algo:LearnSSLSPN}
\end{algorithm}

\subsection{\uppercase{Algorithm}}
The algorithm Maximum Contrastive Pessimistic SPN (MCP-SPN) for learning \safe{} semi-supervised SPNs is illustrated in Algorithm~\ref{algo:LearnSSLSPN} and consists of the following adversarial steps: (1) optimising the \safe{} semi-supervised solution on the given soft labels by maximising a generative or discriminative objective (2) minimising the improvement of the semi-supervised solution over the purely supervised solution by adjusting the soft labels pessimistically. 
As an SPN is a multi-linear function in terms of the model parameters we can apply the generalisation of the minmax theorem for multi-linear functions \cite{Kalantari2016} and interchange the maximisation and the minimisation in our algorithm.

Depending on the choice of the objective, the MCP-SPN procedure first finds a purely supervised solution by only maximising the chosen objective with respect to the labelled data.
Secondly, we initialise all soft labels of the unlabelled data either using an optimistic approach or using random draws from a Dirichlet distribution.
In the case of a generative objective the purely supervised solution can degenerate to a point mass estimator. It is therefore useful for generative SPNs to initialise the soft labels using random draws instead of starting from an optimistic labelling.
After initialising all soft labels the MCP-SPN procedure finds a \safe{} semi-supervised solution $\theta^\ast$ by alternating between the two adversarial steps.
The function call \textit{projectOnSimplex} refers to the approach in \cite{Duchi2008}, which we use to project the soft label assignments back to the $K-1$ simplex (but other approaches for this task could also be used). 
Note that we found it useful to decrease the learning rate $\alpha$ of the pessimistic soft labels adjustment over time. 
In our experiments we therefore used a simple decay function $\alpha \gets \alpha / \sqrt(iteration)$, if necessary more advanced approaches can be used instead.
The source code for \safe{} semi-supervised learning of SPNs is available on-line\footnote{\url{https://github.com/trappmartin/SSLSPN_UAI2017}}.

\section{\uppercase{Experiments}} \label{sec:Experiments}
We analysed the performance of the \safe{} semi-supervised learning approach qualitatively on synthetic data using the generative objective, and quantitatively on various data sets using both objectives.

\subsection{Datasets and Model Generation}
In addition to the synthetic two moons data set \cite{Jain2005}, we used various well known data sets from the UCI repository \cite{Lichman2013} to evaluate the performance of the \safe{} semi-supervised parameter learning approaches. 
We pre-processed the data in the following way: (1) we removed features with zero variance, (2) we applied z-score normalisation.
To ensure broad applicability of the approaches, we selected data sets which origin from a variety of domains and cover a wide range of number of samples and dimensions.
Details on the selected data sets are shown in Table~\ref{tab:datasets} where the last column lists the number of labelled samples used in all experiments.
Note that the number of labelled samples per data set is calculated as in ~\cite{Loog2015}.

To consistently learn SPN structures for all experiments we extended the well-known learnSPN \cite{Gens2013} algorithm for Gaussian distributed data, similar as in \cite{Vergari2015}.
Additionally, we added a layer that conditions on the class labels resulting in structures that are suitable for supervised and semi-supervised learning~\cite{Gens2012}.
As learnSPN produces large SPN structures, which might lead to over-fitting, we used a two step procedure for regularizing the resulting network. 
First, we estimate and apply a pruning depth of the network and secondly, we remove degenerated leaf distributions.
We further ensured throughout all regularization steps that the resulting SPN is complete and decomposable. 

\subsection{\uppercase{Qualitative Results on Synthetic Data}}
\begin{figure*}[t]
\centering
\begin{subfigure}[b]{.3\linewidth}
\includegraphics[width=\linewidth]{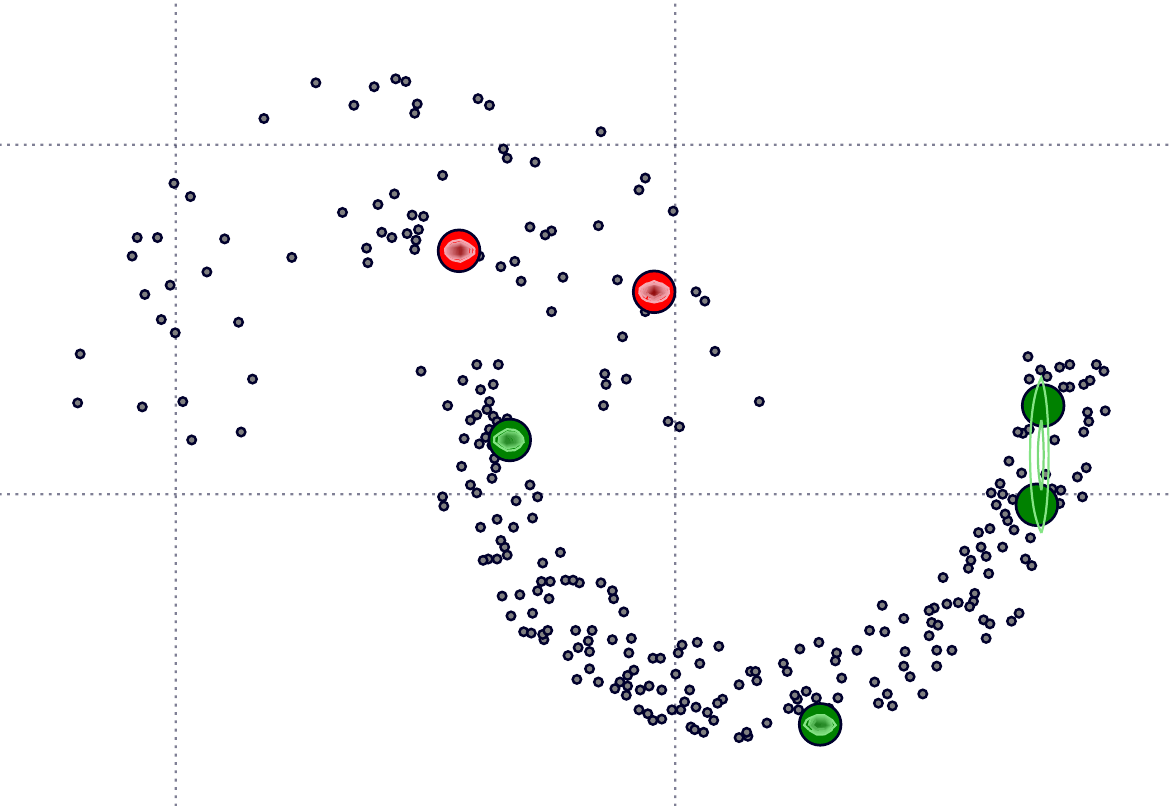}
\caption{Purely Supervised Solution}\label{fig:supervisedJain}
\end{subfigure}
\begin{subfigure}[b]{.3\linewidth}
\includegraphics[width=\linewidth]{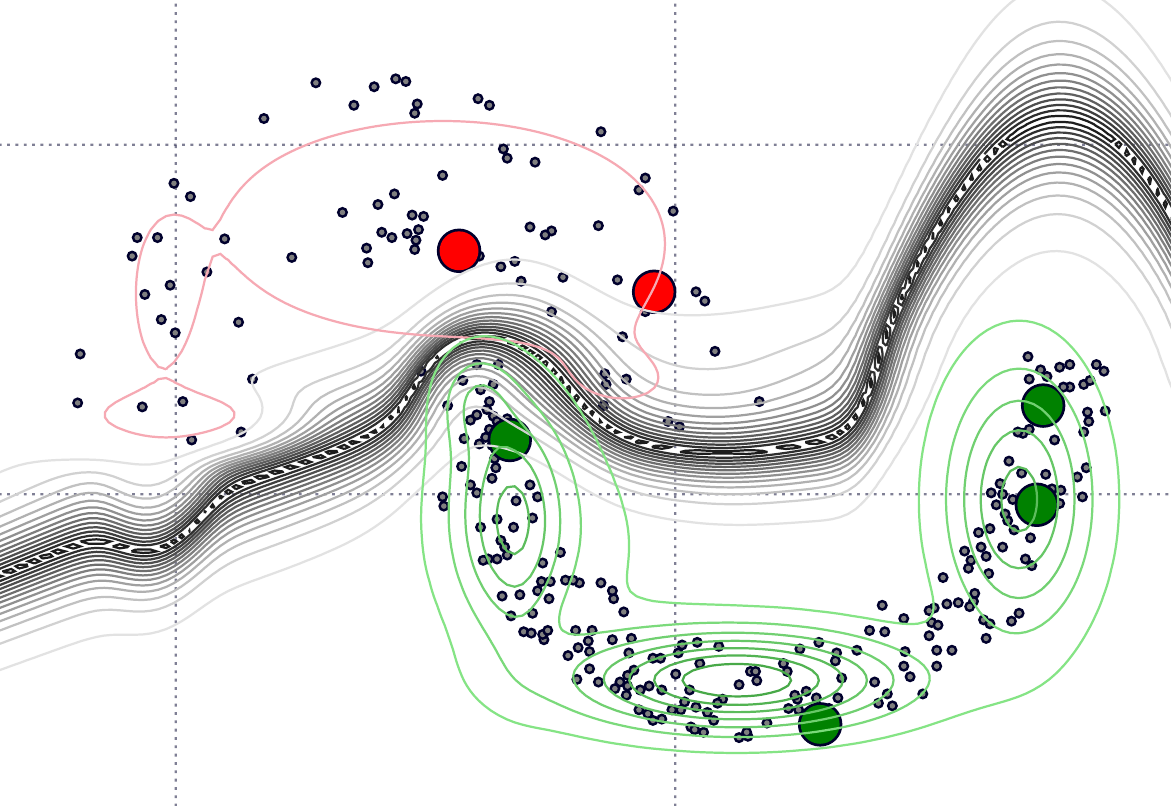}
\caption{Oracle Solution}\label{fig:oracleJain}
\end{subfigure}
\begin{subfigure}[b]{\linewidth}
\begin{adjustbox}{varwidth=\textwidth,fbox,center}
    \begin{subfigure}[b]{.3\linewidth}
    \includegraphics[width=\linewidth]{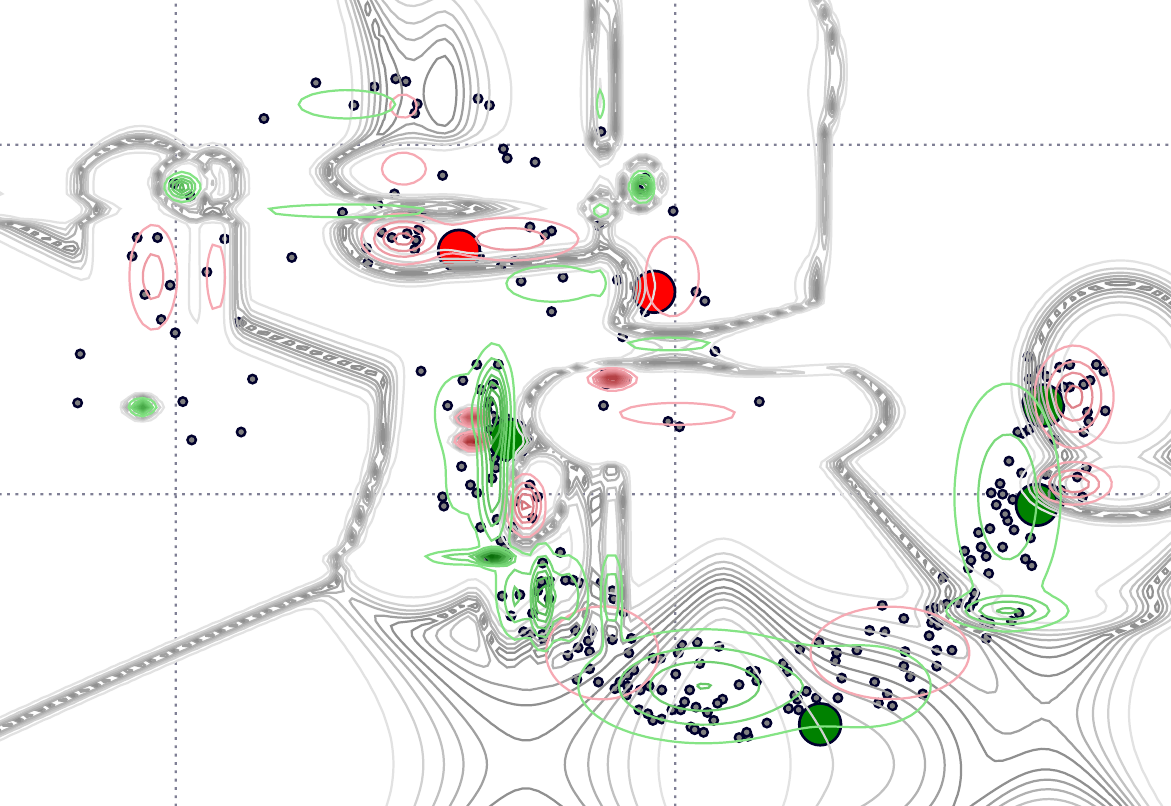}
    \caption*{Iteration 1}\label{fig:sslJain1}
    \end{subfigure}
    \begin{subfigure}[b]{.3\linewidth}
    \includegraphics[width=\linewidth]{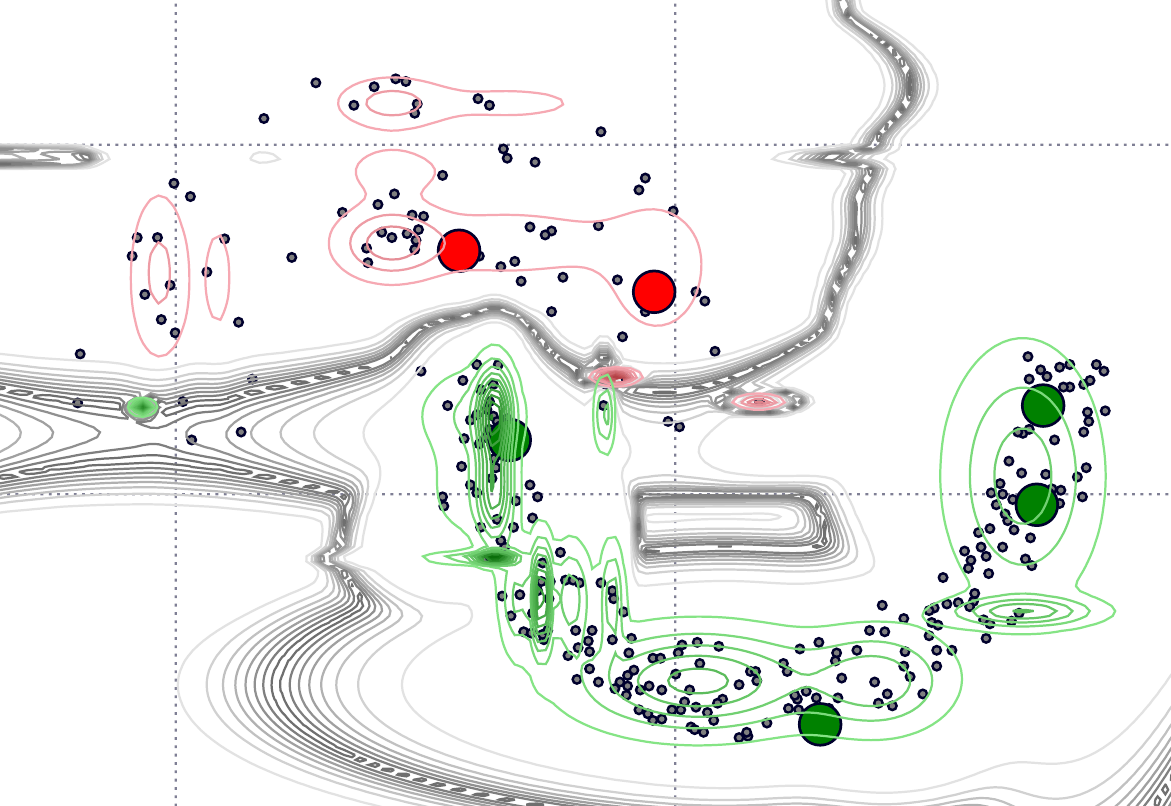}
    \caption*{Iteration 2}\label{fig:sslJain2}
    \end{subfigure}
    \begin{subfigure}[b]{.3\linewidth}
    \includegraphics[width=\linewidth]{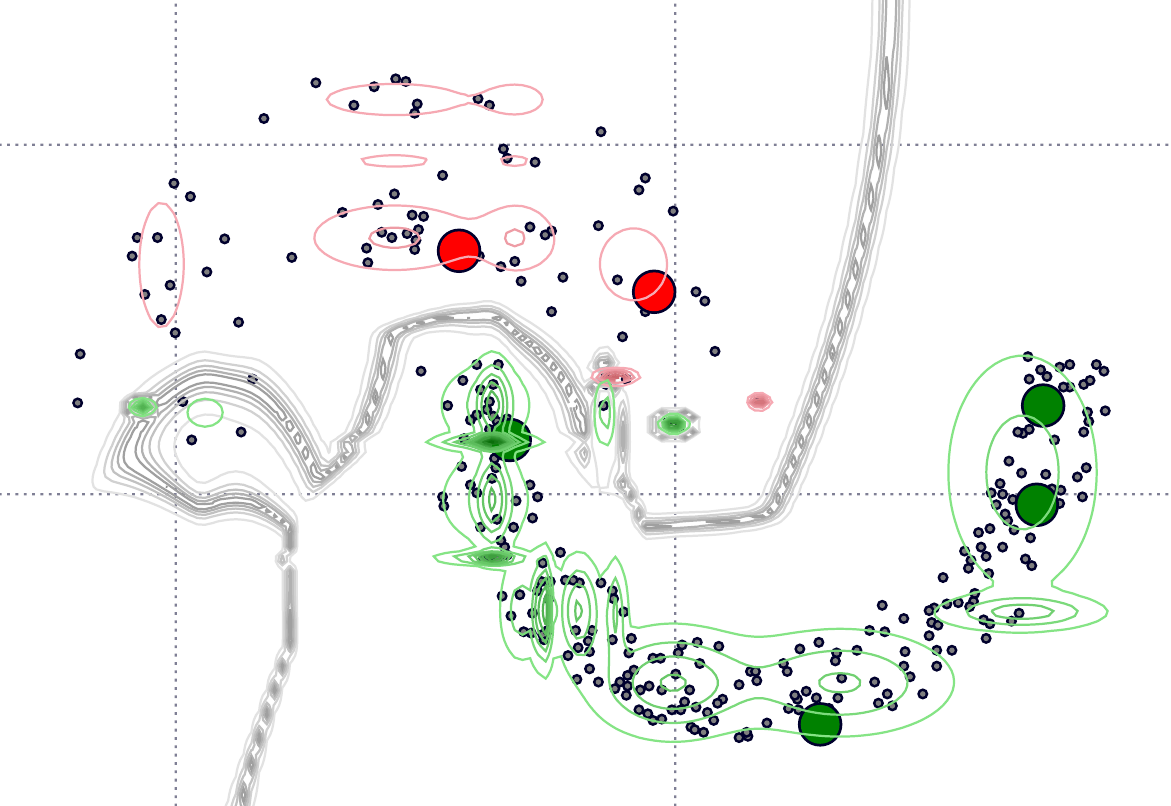}
    \caption*{Converged Model (It. 20)}\label{fig:sslJainFinal}
    \end{subfigure}
\end{adjustbox}
\caption{Safe Semi-Supervised Solution}\label{fig:sslJain}
\end{subfigure}
\caption{Qualitative results on the two moons data set.
The colour of the dots indicate their class label and estimated density regions of the classes are shown using coloured density plots.
Unlabelled samples are shown as small black dots.
The decision boundary of the model is shown using a density plot coloured in grey.
\subref{fig:supervisedJain} Purely supervised solution over-fits the few training examples and degenerates to a point density estimator.
\subref{fig:oracleJain} Oracle solution, 
\subref{fig:sslJain} generative \safe{} semi-supervised parameter learning is able to find a reasonable solution after only a few iterations without making restrictive assumptions on the data distribution.}
\end{figure*}

Due to the non-linearity, flexibility and complexity of SPNs with arbitrary leaf distributions, learning a \safe{} semi-supervised objective for such networks, without enforcing prior assumptions on the data distribution, is much more difficult than for linear models such as Linear Discriminant Analysis (LDA) ~\cite{Loog2015}. 
Therefore, we analysed the behaviour of \safe{} semi-supervised SPNs qualitatively on the synthetic two moons data set \cite{Jain2005}.
Figure~\ref{fig:supervisedJain} shows the purely supervised solution for a small subset of labelled observations and the solution found using a generative \safe{} semi-supervised SPN over time.
For reference the oracle solution, which knows the labels of all observations, is depicted in Figure~\ref{fig:oracleJain}.

\begin{table}
    \centering
\begin{tabular}{@{}lrrrr@{}}
    \toprule
	\textbf{Data Set} & {N} & {D} & {K} & {$2 \cdot D + K$}\\
	\midrule
	BUPA & 345 & 6 & 2 & 14\\
	Fertility & 100 & 9 & 2 & 20\\
	Haberman & 306 & 3 & 2 & 8\\
	ILPD & 583 & 10 & 2 & 22\\
	Ionosphere & 351 & 34 & 2 & 70\\
	Iris & 150 & 4 & 3 & 11\\
	Parkinsons & 197 & 23 & 2 & 48\\
	WDBC & 569 & 32 & 2 & 66\\
	Wine & 178 & 13 & 3 & 29\\
    \bottomrule
\end{tabular}
    \caption{Datasets, details on the number of samples ($N$), dimensionality ($D$), number of classes ($K$) and number of labelled samples used in all evaluations ($2 \cdot D + K$). The number of labelled samples is obtained according to~\cite{Loog2015}.}
    \label{tab:datasets}
\end{table}

The purely supervised SPN clearly over-fits the few labelled examples and degenerated almost completely to a kernel density estimator. 
The \safe{} semi-supervised parameter learning approach is initialised using soft labels drawn from a Dirichlet distribution, to allow the model to escape from the local optimum.
As shown in Figure~\ref{fig:sslJain}, the generative \safe{} semi-supervised approach is able to find a reasonable solution after only three iterations even with a random initialisation of the soft labels.
The model converges after only 20 iterations to a stable solution without enforcing restrictive assumptions on the data distribution.

\subsection{\uppercase{Generative SPN Performance for Safe Semi-Supervised Learning}}
\paragraph{Experimental Setup}
We constructed truncated network structures using learnSPN~\cite{Gens2013}. The truncation levels have been estimated using the Akaike information criterion \cite{Akaike1974}. 
After the structure construction we initialised all soft labels using random draws from a Dirichlet distribution with equal concentration parameter for all classes. 

Furthermore, we lower bounded the variance of the leaf distributions to the $i$th percentile of the nearest neighbour distances of all data points in $\mathcal{X} \cup \mathcal{U}$.
We selected the smallest percentile such that the constructed lower bound is above zero.
Imposing a lower bound on the variances of the leaf node distributions in such way prevents the univariate Gaussian distributions from degeneration with minimal influence on the model expressiveness.

We analysed our approach for generative semi-supervised learning of SPNs by: (1) splitting each dataset into training (80\%) and testing set (20\%), (2) draw $2 \cdot D+K$ labelled samples stratified from each training set as proposed in \cite{Loog2015}. 
We used an additional labelled validation set of $2 \cdot D+K$ samples for early stopping.
In addition to the labelled samples, we used all remaining observations in the training set as unlabelled examples.

\paragraph{Results}
We compare the performance of the \safe{} semi-supervised learning (SSL) approach against the purely supervised solution, an oracle solution and the solution found by the recently introduced inductive approach (MCPLDA)~\cite{Loog2015}.
All models where evaluated on the test set.
The resulting average log likelihood values are estimated over 100 independent runs.
Table~\ref{tab:generativeResults} lists the average log likelihood and the standard errors of all approaches.
Note that the guarantee of the CPLE is on the training set including unlabelled observations. 
We expect however, the performance of the SSL approach on the test set in expectation to be better or similar to the purely supervised learner.

In most cases we could indeed find an improvement of the \safe{} semi-supervised approach over the purely supervised solution. 
In the cases of \textit{Parkinsons}, \textit{WDBC} and \textit{Wine} the purely supervised learner already finds solutions which are close to the oracle solution.
This might be due to the relative simple geometric properties of those data sets.
In this situation, our SSL approach converged to solutions which are close to the purely supervised solution.
In some cases, e.g. \textit{BUPA}, \textit{Fertility}, \textit{Haberman} and \textit{ILPD}, we could find an improvement upon the oracle solution or near oracle solution performance.
Furthermore, \safe{} semi-supervised SPNs generally outperform MCPLDA on almost all data sets in terms of the log likelihood, with one exception being the \textit{Iris} data set.
Moreover, our approach generally reaches very stable results and achieves estimated standard errors lower than those of the supervised and the MCPLDA solution.

\begin{table*}
    \centering
    \begin{tabular}{@{}lr@{\hspace{.2mm}}r@{\hspace{.2mm}}lr@{\hspace{.2mm}}c@{\hspace{.2mm}}lr@{\hspace{.2mm}}r@{\hspace{.2mm}}lr@{\hspace{.2mm}}r@{\hspace{.2mm}}l@{}}
    \toprule
    \textbf{Data Set} & \multicolumn{3}{c}{\textbf{Supervised}} & \multicolumn{3}{c}{\textbf{SSL}} & \multicolumn{3}{c}{\textbf{Oracle}} & \multicolumn{3}{c}{\textbf{MCPLDA}} \\ 
    \midrule
    BUPA & $ \num[round-precision=2,round-mode=places,
           scientific-notation=false]{-438.7523249835144} $ & $\pm$ & $7 \cdot 10^0$ &$ \pmb{\num[round-precision=2,round-mode=places,
           scientific-notation=false]{-7.310053844385056}} $ & $\pm$ & $\num[round-precision=1,round-mode=figures,
          scientific-notation=true, exponent-product = \cdot, retain-zero-exponent = true]{0.0563381084525893} $ &$ \num[round-precision=2,round-mode=places,
           scientific-notation=false]{-8.797429030409612} $ & $\pm$ & $\num[round-precision=1,round-mode=figures,
          scientific-notation=true, exponent-product = \cdot, retain-zero-exponent = true]{0.17549550626857707} $ &$ \num[round-precision=2,round-mode=places,
           scientific-notation=false]{-9.069584242469894} $ & $\pm$ & $\num[round-precision=1,round-mode=figures,
          scientific-notation=true, exponent-product = \cdot, retain-zero-exponent = true]{0.031016501769955664} $\\
    Fertility & $ \num[round-precision=2,round-mode=places,
           scientific-notation=false]{-3.3105646901143744} $ & $\pm$ & $\num[round-precision=1,round-mode=figures,
          scientific-notation=true, exponent-product = \cdot, retain-zero-exponent = true]{0.0317436612903049} $ &$ \pmb{\num[round-precision=2,round-mode=places,
           scientific-notation=false]{-3.0633531726550842}} $ & $\pm$ & $\num[round-precision=1,round-mode=figures,
          scientific-notation=true, exponent-product = \cdot, retain-zero-exponent = true]{0.007318756609840116} $ &$ \num[round-precision=2,round-mode=places,
           scientific-notation=false]{-2.996028176192663} $ & $\pm$ & $\num[round-precision=1,round-mode=figures,
          scientific-notation=true, exponent-product = \cdot, retain-zero-exponent = true]{0.0064064363045572194} $ &$ \num[round-precision=2,round-mode=places,
           scientific-notation=false]{-12.68292961181484} $ & $\pm$ & $\num[round-precision=1,round-mode=figures,
          scientific-notation=true, exponent-product = \cdot, retain-zero-exponent = true]{0.05150164192021896} $\\
    Haberman & $ \num[round-precision=2,round-mode=places,
           scientific-notation=false]{-138.6339601004464} $ & $\pm$ & $4 \cdot 10^0$ &$ \pmb{\num[round-precision=2,round-mode=places,
           scientific-notation=false]{-5.0527888251864965}} $ & $\pm$ & $\num[round-precision=1,round-mode=figures,
          scientific-notation=true, exponent-product = \cdot, retain-zero-exponent = true]{0.06464752655969555} $ &$ \num[round-precision=2,round-mode=places,
           scientific-notation=false]{-5.13813423113883} $ & $\pm$ & $\num[round-precision=1,round-mode=figures,
          scientific-notation=true, exponent-product = \cdot, retain-zero-exponent = true]{0.059533413434605} $ &$ \num[round-precision=2,round-mode=places,
           scientific-notation=false]{-7.827502384091583} $ & $\pm$ & $1 \cdot 10^{-1}$\\
    ILPD & $ \num[round-precision=2,round-mode=places,
           scientific-notation=false]{-5.615035944715549} $ & $\pm$ & $3 \cdot 10^0$ &$ \pmb{\num[round-precision=2,round-mode=places,
           scientific-notation=false]{-1.1487905880874816}} $ & $\pm$ & $\num[round-precision=1,round-mode=figures,
          scientific-notation=true, exponent-product = \cdot, retain-zero-exponent = true]{0.021950362245804333} $ &$ \num[round-precision=2,round-mode=places,
           scientific-notation=false]{-1.0007077300724394} $ & $\pm$ & $\num[round-precision=1,round-mode=figures,
          scientific-notation=true, exponent-product = \cdot, retain-zero-exponent = true]{0.012212524385228981} $ &$ \num[round-precision=2,round-mode=places,
           scientific-notation=false]{-37.539092280603306} $ & $\pm$ & $1 \cdot 10^{-1}$\\
    Ionosphere & $ \num[round-precision=2,round-mode=places,
           scientific-notation=false]{-2.8344732747160957} $ & $\pm$ & $\num[round-precision=1,round-mode=figures,
          scientific-notation=true, exponent-product = \cdot, retain-zero-exponent = true]{0.05086171471974847} $ &$ \pmb{\num[round-precision=2,round-mode=places,
           scientific-notation=false]{-1.608260855118607}} $ & $\pm$ & $\num[round-precision=1,round-mode=figures,
          scientific-notation=true, exponent-product = \cdot, retain-zero-exponent = true]{0.01424083369101005} $ &$ \num[round-precision=2,round-mode=places,
           scientific-notation=false]{-1.5192804336304968} $ & $\pm$ & $\num[round-precision=1,round-mode=figures,
          scientific-notation=true, exponent-product = \cdot, retain-zero-exponent = true]{0.009128697123122856} $ &$ \num[round-precision=2,round-mode=places,
           scientific-notation=false]{-46.1167983098879} $ & $\pm$ & $\num[round-precision=1,round-mode=figures,
          scientific-notation=true, exponent-product = \cdot, retain-zero-exponent = true]{0.05382252702983977} $\\
    Iris & $ \num[round-precision=2,round-mode=places,
           scientific-notation=false]{-20.648365591948483} $ & $\pm$ & $\num[round-precision=1,round-mode=figures,
          scientific-notation=true, exponent-product = \cdot, retain-zero-exponent = true]{0.8508286568270595} $ &$ \num[round-precision=2,round-mode=places,
           scientific-notation=false]{-3.7780487990911364} $ & $\pm$ & $\num[round-precision=1,round-mode=figures,
          scientific-notation=true, exponent-product = \cdot, retain-zero-exponent = true]{0.03307402843175255} $ &$ \num[round-precision=2,round-mode=places,
           scientific-notation=false]{-2.1668453853252663} $ & $\pm$ & $\num[round-precision=1,round-mode=figures,
          scientific-notation=true, exponent-product = \cdot, retain-zero-exponent = true]{0.013701046855750762} $ &$ \pmb{\num[round-precision=2,round-mode=places,
           scientific-notation=false]{-2.6450502648208487}} $ & $\pm$ & $\num[round-precision=1,round-mode=figures,
          scientific-notation=true, exponent-product = \cdot, retain-zero-exponent = true]{0.046838185088918585} $\\
    Parkinsons & $ \pmb{\num[round-precision=2,round-mode=places,
           scientific-notation=false]{-1.324222726610557}} $ & $\pm$ & $\num[round-precision=1,round-mode=figures,
          scientific-notation=true, exponent-product = \cdot, retain-zero-exponent = true]{0.004456104961874492} $ &$ \num[round-precision=2,round-mode=places,
           scientific-notation=false]{-1.3423741998275522} $ & $\pm$ & $\num[round-precision=1,round-mode=figures,
          scientific-notation=true, exponent-product = \cdot, retain-zero-exponent = true]{0.004083711259148531} $ &$ \num[round-precision=2,round-mode=places,
           scientific-notation=false]{-1.2975783074242437} $ & $\pm$ & $\num[round-precision=1,round-mode=figures,
          scientific-notation=true, exponent-product = \cdot, retain-zero-exponent = true]{0.002296871583123938} $ &$ \num[round-precision=2,round-mode=places,
           scientific-notation=false]{-2.2693472712741016} $ & $\pm$ & $\num[round-precision=1,round-mode=figures,
          scientific-notation=true, exponent-product = \cdot, retain-zero-exponent = true]{0.04895012224871527} $\\
    WDBC & $ \pmb{\num[round-precision=2,round-mode=places,
           scientific-notation=false]{-1.8971513587222495}} $ & $\pm$ & $\num[round-precision=1,round-mode=figures,
          scientific-notation=true, exponent-product = \cdot, retain-zero-exponent = true]{0.001389003531740278} $ &$ \num[round-precision=2,round-mode=places,
           scientific-notation=false]{-1.9263515531393403} $ & $\pm$ & $\num[round-precision=1,round-mode=figures,
          scientific-notation=true, exponent-product = \cdot, retain-zero-exponent = true]{0.0016043885628163924} $ &$ \num[round-precision=2,round-mode=places,
           scientific-notation=false]{-1.882955591550101} $ & $\pm$ & $\num[round-precision=1,round-mode=figures,
          scientific-notation=true, exponent-product = \cdot, retain-zero-exponent = true]{0.00032648611354224745} $ &$ \num[round-precision=2,round-mode=places,
           scientific-notation=false]{-10.753032946846195} $ & $\pm$ & $\num[round-precision=1,round-mode=figures,
          scientific-notation=true, exponent-product = \cdot, retain-zero-exponent = true]{0.012177344919801377} $\\
    Wine & $ \pmb{\num[round-precision=2,round-mode=places,
           scientific-notation=false]{-2.46928960091736}} $ & $\pm$ & $\num[round-precision=1,round-mode=figures,
          scientific-notation=true, exponent-product = \cdot, retain-zero-exponent = true]{0.004022816579402819} $ &$ \pmb{\num[round-precision=2,round-mode=places,
           scientific-notation=false]{-2.4713558531391906}} $ & $\pm$ & $\num[round-precision=1,round-mode=figures,
          scientific-notation=true, exponent-product = \cdot, retain-zero-exponent = true]{0.002448648863526557} $ &$ \num[round-precision=2,round-mode=places,
           scientific-notation=false]{-2.437730088143566} $ & $\pm$ & $\num[round-precision=1,round-mode=figures,
          scientific-notation=true, exponent-product = \cdot, retain-zero-exponent = true]{0.0009130644377289696} $ &$ \num[round-precision=2,round-mode=places,
           scientific-notation=false]{-15.278947136892224} $ & $\pm$ & $\num[round-precision=1,round-mode=figures,
          scientific-notation=true, exponent-product = \cdot, retain-zero-exponent = true]{0.019953374907304564} $\\
    \bottomrule
    \end{tabular}
    \caption{Averaged log likelihood and standard errors estimated on the test set over 100 independent trials. The best results for each data set obtained by a supervised or semi-supervised model are shown in \textbf{bold} face.}
    \label{tab:generativeResults}
\end{table*}

\subsection{\uppercase{Discriminative SPN Performance for Safe Semi-Supervised Learning}}
We assess the classification performance of discriminative \safe{} semi-supervised learning below, as optimising a discriminative objective is a more natural way for classification tasks. 

\paragraph{Experimental Setup}
Similar to the quantitative evaluation of the generative approach, we constructed truncated structures for all experiments.
To avoid over-fitting we used early truncation of the model, estimated according to the performance on the validation set.
We further initialised all soft labels using optimistic predictions from the purely supervised model.
To obtain training and test sets, we followed the same approach as described for the generative experiments.
Similar to the generative evaluation, the randomly drawn labelled subset is obtained from the training set and the performance of each algorithm is estimated over 100 independent trials.

\paragraph{Results}
We compared the performance of our discriminative approach against the purely supervised solution, the oracle solution and the following state of the art approaches: Transductive SVM (TSVM)~\cite{Collobert2006}, Minimum Entropy Regularization (MER)~\cite{Grandvalet2004} and the recently published Implicitly Constrained Least Squares (ICLS)~\cite{Krijthe2015}.
To assess the performance of a classification method, we computed the $F_1$ score for binary classification tasks.
In cases of multi-class data sets, we used the macro average $F_1$ score.
To compute multi-class predictions for approaches designed only for binary classification we used the one-vs-rest approach.
The average $F_1$ scores as well as the standard errors of all approaches are shown in Table~\ref{tab:discriminativeResults}.

The \safe{} semi-supervised parameter learning approach achieves competitive results for almost all data sets. 
In general, our approach produces reasonable results and does not degenerate if certain assumptions are not met.
Moreover, in several cases our discriminative approach achieves test $F_1$ scores which are comparable to those of the oracle solution, e.g. for \textit{Haberman} and \textit{Wine}.
We could find the lowest performance of our approach on the \textit{Fertility} data set.
Note that the $F_1$ scores on \textit{Fertility}, \textit{Haberman} and \textit{ILPD} are generally very low as those are imbalanced or skewed data sets.

In general, the proposed \safe{} semi-supervised learning for SPNs is a powerful adversarial approach which scales linearly in the number of samples and is non-restrictive.
Even though we achieved competitive results even on data sets where low density assumptions are met, e.g. \textit{Wine}, further improvements may be achieved by trading off optimism and pessimism.
One way of approaching this issue would be to add a weighting scheme into the CPLE formulation.

Even though optimising the conditional log likelihood inside the CPLE objective provides a reasonable criterion for classification tasks, this approach does not guarantee to improve the classification performance of the learner.
It is therefore possible, that better classification performance can be achieved by using a multi-class squared-hinge loss, which was recently used in a related model~\cite{Gens2017}.

\section{\uppercase{Conclusion and Future Work}} \label{sec:Conclusion}
In this paper, we introduced the first approach for semi-supervised parameter learning with Sum-Product Networks (SPNs).
We presented generative and discriminative \safe{} semi-supervised learning procedures which guarantee that adding unlabelled data can increase, but not degrade, the performance of the learner on the training set.
Furthermore, our approach exploits the tractability of SPNs and scales linear in the number of data points and model parameters.
In contrast to other semi-supervised learners, the proposed approach is non-restrictive and does not need prior assumptions on the data distribution.
The approach allows broad applicability and is a generic \safe{} semi-supervised learning procedure for all models which leverage the sum-product theorem \cite{Friesen2016} and therefore provides a semi-supervised learning procedure beyond SPNs.

\begin{table*}
\centering
\begin{tabular}{@{}lr@{\hspace{.2mm}}r@{\hspace{.2mm}}lr@{\hspace{.2mm}}c@{\hspace{.2mm}}lr@{\hspace{.2mm}}r@{\hspace{.2mm}}lr@{\hspace{.2mm}}r@{\hspace{.2mm}}lr@{\hspace{.2mm}}r@{\hspace{.2mm}}lr@{\hspace{.2mm}}r@{\hspace{.2mm}}l@{}}
\toprule
\textbf{Data Set} & \multicolumn{3}{c}{\textbf{Supervised}} & \multicolumn{3}{c}{\textbf{SSL}} & \multicolumn{3}{c}{\textbf{Oracle}} & \multicolumn{3}{c}{\textbf{TSVM}} & \multicolumn{3}{c}{\textbf{ICLSC}} & \multicolumn{3}{c}{\textbf{MER}} \\ 
\midrule
BUPA & $ \num[round-precision=2,round-mode=places,
           scientific-notation=false]{0.4132066976761657} $ & $\pm$ & $\num[round-precision=1,round-mode=figures,
          scientific-notation=true, exponent-product = \cdot, retain-zero-exponent = true]{0.012329166668098928} $ &$ \num[round-precision=2,round-mode=places,
           scientific-notation=false]{0.40109400007927226} $ & $\pm$ & $\num[round-precision=1,round-mode=figures,
          scientific-notation=true, exponent-product = \cdot, retain-zero-exponent = true]{0.012415181227102591} $ &$ \num[round-precision=2,round-mode=places,
           scientific-notation=false]{0.4787564808611785} $ & $\pm$ & $\num[round-precision=1,round-mode=figures,
          scientific-notation=true, exponent-product = \cdot, retain-zero-exponent = true]{0.004608613570946731} $ &$ \num[round-precision=2,round-mode=places,
           scientific-notation=false]{0.36193833421494837} $ & $\pm$ & $\num[round-precision=1,round-mode=figures,
          scientific-notation=true, exponent-product = \cdot, retain-zero-exponent = true]{0.0163806598297098} $ &$ \pmb{\num[round-precision=2,round-mode=places,
           scientific-notation=false]{0.4724080498005464}} $ & $\pm$ & $\num[round-precision=1,round-mode=figures,
          scientific-notation=true, exponent-product = \cdot, retain-zero-exponent = true]{0.007089063543091236} $ &$ \num[round-precision=2,round-mode=places,
           scientific-notation=false]{0.42001893438437465} $ & $\pm$ & $\num[round-precision=1,round-mode=figures,
          scientific-notation=true, exponent-product = \cdot, retain-zero-exponent = true]{0.013773974428820338} $\\
Fertility & $ \num[round-precision=2,round-mode=places,
           scientific-notation=false]{0.07408802308802308} $ & $\pm$ & $\num[round-precision=1,round-mode=figures,
          scientific-notation=true, exponent-product = \cdot, retain-zero-exponent = true]{0.016912532946779624} $ &$ \num[round-precision=2,round-mode=places,
           scientific-notation=false]{0.029333333333333333} $ & $\pm$ & $\num[round-precision=1,round-mode=figures,
          scientific-notation=true, exponent-product = \cdot, retain-zero-exponent = true]{0.012320814467225877} $ &$ \num[round-precision=2,round-mode=places,
           scientific-notation=false]{0.05854761904761904} $ & $\pm$ & $\num[round-precision=1,round-mode=figures,
          scientific-notation=true, exponent-product = \cdot, retain-zero-exponent = true]{0.016072523077475982} $ &$ \num[round-precision=2,round-mode=places,
           scientific-notation=false]{0.0677936507936508} $ & $\pm$ & $\num[round-precision=1,round-mode=figures,
          scientific-notation=true, exponent-product = \cdot, retain-zero-exponent = true]{0.01592002490634426} $ &$ \num[round-precision=2,round-mode=places,
           scientific-notation=false]{0.07321428571428572} $ & $\pm$ & $\num[round-precision=1,round-mode=figures,
          scientific-notation=true, exponent-product = \cdot, retain-zero-exponent = true]{0.0177770168036989} $ &$ \pmb{\num[round-precision=2,round-mode=places,
           scientific-notation=false]{0.12387373737373736}} $ & $\pm$ & $\num[round-precision=1,round-mode=figures,
          scientific-notation=true, exponent-product = \cdot, retain-zero-exponent = true]{0.01905710079186234} $\\
Haber. & $ \num[round-precision=2,round-mode=places,
           scientific-notation=false]{0.23445201381613842} $ & $\pm$ & $\num[round-precision=1,round-mode=figures,
          scientific-notation=true, exponent-product = \cdot, retain-zero-exponent = true]{0.0186607146016289} $ &$ \num[round-precision=2,round-mode=places,
           scientific-notation=false]{0.2819689409884862} $ & $\pm$ & $\num[round-precision=1,round-mode=figures,
          scientific-notation=true, exponent-product = \cdot, retain-zero-exponent = true]{0.015906023937611632} $ &$ \num[round-precision=2,round-mode=places,
           scientific-notation=false]{0.25} $ & $\pm$ & $\num[round-precision=1,round-mode=figures,
          scientific-notation=true, exponent-product = \cdot, retain-zero-exponent = true]{0.0} $ &$ \num[round-precision=2,round-mode=places,
           scientific-notation=false]{0.19517484091608533} $ & $\pm$ & $\num[round-precision=1,round-mode=figures,
          scientific-notation=true, exponent-product = \cdot, retain-zero-exponent = true]{0.017654320725359652} $ &$ \pmb{\num[round-precision=2,round-mode=places,
           scientific-notation=false]{0.3264368360160646}} $ & $\pm$ & $\num[round-precision=1,round-mode=figures,
          scientific-notation=true, exponent-product = \cdot, retain-zero-exponent = true]{0.013158621413238782} $ &$ \num[round-precision=2,round-mode=places,
           scientific-notation=false]{0.2660651515298579} $ & $\pm$ & $\num[round-precision=1,round-mode=figures,
          scientific-notation=true, exponent-product = \cdot, retain-zero-exponent = true]{0.01455753643171881} $\\
ILPD & $ \num[round-precision=2,round-mode=places,
           scientific-notation=false]{0.17141063641708823} $ & $\pm$ & $\num[round-precision=1,round-mode=figures,
          scientific-notation=true, exponent-product = \cdot, retain-zero-exponent = true]{0.01585717915749354} $ &$ \num[round-precision=2,round-mode=places,
           scientific-notation=false]{0.19867766780631727} $ & $\pm$ & $\num[round-precision=1,round-mode=figures,
          scientific-notation=true, exponent-product = \cdot, retain-zero-exponent = true]{0.014872842713238738} $ &$ \num[round-precision=2,round-mode=places,
           scientific-notation=false]{0.24446147212844715} $ & $\pm$ & $\num[round-precision=1,round-mode=figures,
          scientific-notation=true, exponent-product = \cdot, retain-zero-exponent = true]{0.003780770723645524} $ &$ \num[round-precision=2,round-mode=places,
           scientific-notation=false]{0.22561018106462727} $ & $\pm$ & $\num[round-precision=1,round-mode=figures,
          scientific-notation=true, exponent-product = \cdot, retain-zero-exponent = true]{0.01841064762520556} $ &$ \num[round-precision=2,round-mode=places,
           scientific-notation=false]{0.28553799160779303} $ & $\pm$ & $\num[round-precision=1,round-mode=figures,
          scientific-notation=true, exponent-product = \cdot, retain-zero-exponent = true]{0.010488585944443036} $ &$ \pmb{\num[round-precision=2,round-mode=places,
           scientific-notation=false]{0.3272334434201049}} $ & $\pm$ & $\num[round-precision=1,round-mode=figures,
          scientific-notation=true, exponent-product = \cdot, retain-zero-exponent = true]{0.01648579913640548} $\\
Ionos. & $ \num[round-precision=2,round-mode=places,
           scientific-notation=false]{0.7935741162730914} $ & $\pm$ & $\num[round-precision=1,round-mode=figures,
          scientific-notation=true, exponent-product = \cdot, retain-zero-exponent = true]{0.00412513316874567} $ &$ \pmb{\num[round-precision=2,round-mode=places,
           scientific-notation=false]{0.8221724829195172}} $ & $\pm$ & $\num[round-precision=1,round-mode=figures,
          scientific-notation=true, exponent-product = \cdot, retain-zero-exponent = true]{0.0038602590192728} $ &$ \num[round-precision=2,round-mode=places,
           scientific-notation=false]{0.8695652173913043} $ & $\pm$ & $\num[round-precision=1,round-mode=figures,
          scientific-notation=true, exponent-product = \cdot, retain-zero-exponent = true]{0.0} $ &$ \num[round-precision=2,round-mode=places,
           scientific-notation=false]{0.6605037794236562} $ & $\pm$ & $\num[round-precision=1,round-mode=figures,
          scientific-notation=true, exponent-product = \cdot, retain-zero-exponent = true]{0.008658182401680831} $ &$ \num[round-precision=2,round-mode=places,
           scientific-notation=false]{0.6118653075874538} $ & $\pm$ & $\num[round-precision=1,round-mode=figures,
          scientific-notation=true, exponent-product = \cdot, retain-zero-exponent = true]{0.008521554066407314} $ &$ \num[round-precision=2,round-mode=places,
           scientific-notation=false]{0.7000051755343017} $ & $\pm$ & $\num[round-precision=1,round-mode=figures,
          scientific-notation=true, exponent-product = \cdot, retain-zero-exponent = true]{0.007391904199019017} $\\
Iris & $ \num[round-precision=2,round-mode=places,
           scientific-notation=false]{0.7285948373821634} $ & $\pm$ & $\num[round-precision=1,round-mode=figures,
          scientific-notation=true, exponent-product = \cdot, retain-zero-exponent = true]{0.01218831603599306} $ &$ \pmb{\num[round-precision=2,round-mode=places,
           scientific-notation=false]{0.8819632810831786}} $ & $\pm$ & $\num[round-precision=1,round-mode=figures,
          scientific-notation=true, exponent-product = \cdot, retain-zero-exponent = true]{0.012302731951361811} $ &$ \num[round-precision=2,round-mode=places,
           scientific-notation=false]{0.932659932659933} $ & $\pm$ & $\num[round-precision=1,round-mode=figures,
          scientific-notation=true, exponent-product = \cdot, retain-zero-exponent = true]{0.} $ &$ \num[round-precision=2,round-mode=places,
           scientific-notation=false]{0.7198578913257322} $ & $\pm$ & $\num[round-precision=1,round-mode=figures,
          scientific-notation=true, exponent-product = \cdot, retain-zero-exponent = true]{0.01272710494458383} $ &$ \num[round-precision=2,round-mode=places,
           scientific-notation=false]{0.7437223281902567} $ & $\pm$ & $\num[round-precision=1,round-mode=figures,
          scientific-notation=true, exponent-product = \cdot, retain-zero-exponent = true]{0.0184956376362544} $ &$ \num[round-precision=2,round-mode=places,
           scientific-notation=false]{0.8047297259746781} $ & $\pm$ & $\num[round-precision=1,round-mode=figures,
          scientific-notation=true, exponent-product = \cdot, retain-zero-exponent = true]{0.006225744259213518} $\\
Parkins. & $ \num[round-precision=2,round-mode=places,
           scientific-notation=false]{0.7197686670055092} $ & $\pm$ & $\num[round-precision=1,round-mode=figures,
          scientific-notation=true, exponent-product = \cdot, retain-zero-exponent = true]{0.010336458724162445} $ &$ \pmb{\num[round-precision=2,round-mode=places,
           scientific-notation=false]{0.7742401413515967}} $ & $\pm$ & $\num[round-precision=1,round-mode=figures,
          scientific-notation=true, exponent-product = \cdot, retain-zero-exponent = true]{0.00432241852354864} $ &$ \num[round-precision=2,round-mode=places,
           scientific-notation=false]{0.8199095022624437} $ & $\pm$ & $\num[round-precision=1,round-mode=figures,
          scientific-notation=true, exponent-product = \cdot, retain-zero-exponent = true]{0.0036199095022624423} $ &$ \num[round-precision=2,round-mode=places,
           scientific-notation=false]{0.7367271115460174} $ & $\pm$ & $\num[round-precision=1,round-mode=figures,
          scientific-notation=true, exponent-product = \cdot, retain-zero-exponent = true]{0.01099897275094066} $ &$ \num[round-precision=2,round-mode=places,
           scientific-notation=false]{0.6645965952124184} $ & $\pm$ & $\num[round-precision=1,round-mode=figures,
          scientific-notation=true, exponent-product = \cdot, retain-zero-exponent = true]{0.01551381825766443} $ &$ \num[round-precision=2,round-mode=places,
           scientific-notation=false]{0.6793774815259672} $ & $\pm$ & $\num[round-precision=1,round-mode=figures,
          scientific-notation=true, exponent-product = \cdot, retain-zero-exponent = true]{0.013807839764549726} $\\
PID & $ \num[round-precision=2,round-mode=places,
           scientific-notation=false]{0.3806789058616145} $ & $\pm$ & $\num[round-precision=1,round-mode=figures,
          scientific-notation=true, exponent-product = \cdot, retain-zero-exponent = true]{0.013585852845410648} $ &$ \num[round-precision=2,round-mode=places,
           scientific-notation=false]{0.44994900524849935} $ & $\pm$ & $\num[round-precision=1,round-mode=figures,
          scientific-notation=true, exponent-product = \cdot, retain-zero-exponent = true]{0.010660561390133851} $ &$ \num[round-precision=2,round-mode=places,
           scientific-notation=false]{0.6389878447872835} $ & $\pm$ & $\num[round-precision=1,round-mode=figures,
          scientific-notation=true, exponent-product = \cdot, retain-zero-exponent = true]{0.0007502020329979597} $ &$ \num[round-precision=2,round-mode=places,
           scientific-notation=false]{0.4547244433065946} $ & $\pm$ & $\num[round-precision=1,round-mode=figures,
          scientific-notation=true, exponent-product = \cdot, retain-zero-exponent = true]{0.013858261072043637} $ &$ \num[round-precision=2,round-mode=places,
           scientific-notation=false]{0.5426311639175784} $ & $\pm$ & $\num[round-precision=1,round-mode=figures,
          scientific-notation=true, exponent-product = \cdot, retain-zero-exponent = true]{0.007203105925293758} $ &$ \pmb{\num[round-precision=2,round-mode=places,
           scientific-notation=false]{0.5652028555319386}} $ & $\pm$ & $\num[round-precision=1,round-mode=figures,
          scientific-notation=true, exponent-product = \cdot, retain-zero-exponent = true]{0.008936565121020432} $\\
WDBC & $ \num[round-precision=2,round-mode=places,
           scientific-notation=false]{0.8508528208374364} $ & $\pm$ & $\num[round-precision=1,round-mode=figures,
          scientific-notation=true, exponent-product = \cdot, retain-zero-exponent = true]{0.002870271159614409} $ &$ \num[round-precision=2,round-mode=places,
           scientific-notation=false]{0.8953026733310299} $ & $\pm$ & $\num[round-precision=1,round-mode=figures,
          scientific-notation=true, exponent-product = \cdot, retain-zero-exponent = true]{0.0024561395370101507} $ &$ \num[round-precision=2,round-mode=places,
           scientific-notation=false]{0.9225138019874862} $ & $\pm$ & $\num[round-precision=1,round-mode=figures,
          scientific-notation=true, exponent-product = \cdot, retain-zero-exponent = true]{0.00034269679560054776} $ &$ \num[round-precision=2,round-mode=places,
           scientific-notation=false]{0.9113968764228546} $ & $\pm$ & $\num[round-precision=1,round-mode=figures,
          scientific-notation=true, exponent-product = \cdot, retain-zero-exponent = true]{0.004274473073404662} $ &$ \num[round-precision=2,round-mode=places,
           scientific-notation=false]{0.8822056075537128} $ & $\pm$ & $\num[round-precision=1,round-mode=figures,
          scientific-notation=true, exponent-product = \cdot, retain-zero-exponent = true]{0.00394755609541206} $ &$ \pmb{\num[round-precision=2,round-mode=places,
           scientific-notation=false]{0.9160225087044541}} $ & $\pm$ & $\num[round-precision=1,round-mode=figures,
          scientific-notation=true, exponent-product = \cdot, retain-zero-exponent = true]{0.0032816212610284915} $\\
Wine & $ \num[round-precision=2,round-mode=places,
           scientific-notation=false]{0.8183306463033395} $ & $\pm$ & $\num[round-precision=1,round-mode=figures,
          scientific-notation=true, exponent-product = \cdot, retain-zero-exponent = true]{0.006775190641706283} $ &$ \pmb{\num[round-precision=2,round-mode=places,
           scientific-notation=false]{0.9744300255633804}} $ & $\pm$ & $\num[round-precision=1,round-mode=figures,
          scientific-notation=true, exponent-product = \cdot, retain-zero-exponent = true]{0.00242382774444278} $ &$ \num[round-precision=2,round-mode=places,
           scientific-notation=false]{0.9694643527536567} $ & $\pm$ & $\num[round-precision=1,round-mode=figures,
          scientific-notation=true, exponent-product = \cdot, retain-zero-exponent = true]{0.0035855604106603523} $ &$ \num[round-precision=2,round-mode=places,
           scientific-notation=false]{0.9646158430546294} $ & $\pm$ & $\num[round-precision=1,round-mode=figures,
          scientific-notation=true, exponent-product = \cdot, retain-zero-exponent = true]{0.0022155764053527214} $ &$ \num[round-precision=2,round-mode=places,
           scientific-notation=false]{0.9534946376076071} $ & $\pm$ & $\num[round-precision=1,round-mode=figures,
          scientific-notation=true, exponent-product = \cdot, retain-zero-exponent = true]{0.006547276169471291} $ &$ \num[round-precision=2,round-mode=places,
           scientific-notation=false]{0.9467653332682499} $ & $\pm$ & $\num[round-precision=1,round-mode=figures,
          scientific-notation=true, exponent-product = \cdot, retain-zero-exponent = true]{0.008525134330505686} $\\
\bottomrule
\end{tabular}
\caption{Macro-average F1 scores estimated on the test set over 100 independent trials. The best results for each data set obtained by a supervised or semi-supervised model are shown in \textbf{bold} face.}
\label{tab:discriminativeResults}
\end{table*}

We investigated the performance of our approach quantitatively and qualitatively.
In the conducted qualitative analysis we found that the generative \safe{} semi-supervised parameter learning approach is able to a find reasonable solution after only a few iterations and is able to escape from the degenerated supervised solutions.
We further compared the performance of \safe{} semi-supervised parameter learning for SPNs against state-of-the-art approaches.
The proposed \safe{} semi-supervised learning for SPNs achieves competitive performance compared to state-of-the-art approaches, and outperformed supervised SPNs in the majority of cases.
Even though our approach is non-restrictive and does not need prior assumptions on the data distribution, \safe{} semi-supervised SPNs can utilise low density regions if the structure of the network reflects geometric properties of the data distribution.
However, as such assumptions are not enforced in the learning procedure, our \safe{} semi-supervised learner is still capable of finding decision boundaries which cross high density regions.

Future research directions include: interleaving network structure learning with semi-supervised parameter learning, extensions to other learning objectives, investigating possibilities for trading off optimism and pessimism in the objective, dealing with covariate shift and analysing instability in \safe{} semi-supervised SPNs and its comparison with GANs.
Furthermore, we plan to apply our \safe{} semi-supervised learning approach to high-dimensional classification problems from medicine, genetics and other domains.

\section*{Acknowledgments}
This research is partially funded by the Austrian Science Fund (FWF): P 27530 and P 27803-N15.

\bibliographystyle{plain}
\bibliography{SumProductNetworks,SemiSupervisedLearning,Misc}

\end{document}